# From safe screening rules to working sets for faster Lasso-type solvers

Mathurin Massias[*]  Alexandre Gramfort[*]  Joseph Salmon[*]

May 1, 2017

## Abstract

Convex sparsity-promoting regularizations are ubiquitous in modern statistical learning. By construction, they yield solutions with few non-zero coefficients, which correspond to saturated constraints in the dual optimization formulation. Working set (WS) strategies are generic optimization techniques that consist in solving simpler problems that only consider a subset of constraints, whose indices form the WS. Working set methods therefore involve two nested iterations: the outer loop corresponds to the definition of the WS and the inner loop calls a solver for the subproblems. For the Lasso estimator a WS is a set of features, while for a Group Lasso it refers to a set of groups. In practice, WS are generally small in this context so the associated feature Gram matrix can fit in memory. Here we show that the Gauss-Southwell rule (a greedy strategy for block coordinate descent techniques) leads to fast solvers in this case. Combined with a working set strategy based on an aggressive use of so-called Gap Safe screening rules, we propose a solver achieving state-of-the-art performance on sparse learning problems. Results are presented on Lasso and multi-task Lasso estimators.

## 1 Introduction

Sparsity-promoting regularization has had a considerable impact on high dimensional statistics both in terms of applications and on the theoretical side (with finite sample results as well as asymptotic ones involving exponentially more features than the underlying sparsity (Bickel et al., 2009). Yet it comes with a cost, as inferring parameters for such sparse estimators requires solving high-dimensional constrained or non-smooth optimization problems, for which dedicated advanced solvers are necessary (Bach et al., 2012).

While sparsity can come to the rescue of statistical theory, it can also be exploited to come up with faster solvers. Various optimization strategies have been proposed to accelerate the solvers for problems such as Lasso or sparse logistic regression involving $\ell_1$ regularization, multi-task Lasso, multinomial logistic or group-Lasso involving $\ell_1/\ell_2$ mixed-norms (Osborne et al., 2000; Koh et al., 2007; Friedman et al., 2007). We will refer to these problems as Lasso-type problems (Bach et al., 2012). For statistical machine learning, as opposed to fields such as signal processing which often involve implicit operators (*e.g.*, FFTs, wavelets), design matrices, which store feature values, are explicit sparse or dense matrices. For Lasso-type problems, this fact has led to the massive success of so-called (block) coordinate descent (BCD) techniques (Tseng, 2001; Friedman et al., 2007; Wu and Lange, 2008; Shalev-Shwartz and Zhang, 2016), which consist in updating one coordinate or one block of coordinates at a time. Different BCD strategies exist depending on how one iterates over coordinates: it can be a cyclic rule as used by Friedman et al. (2007), random (Shalev-Shwartz and Zhang, 2016), or greedy (meaning that the coordinate which is updated is the one leading to the best improvement on the objective or on a surrogate (Shevade and Keerthi, 2003; Wu and Lange, 2008)). The later rule, recently studied by Tseng and Yun (2009); Nutini et al. (2015); Peng et al. (2016) is historically known as the Gauss-Southwell (GS) rule (Southwell, 1941).

[*]LTCI, Télécom ParisTech, Université Paris-Saclay, 46 rue Barrault, 75013, Paris, France `first.last@telecom-paristech.fr`

To further scale up generic solvers, one recurrent idea in machine learning has been to limit the size of the problems solved. This has been popularized for SVM (Joachims, 1998; Fan et al., 2008), but is also a natural strategy for the Lasso as the solution is expected to be sparse. This idea is at the heart of the so-called *strong rules* (Tibshirani et al., 2012) used in the popular GLMNET R package, but similar ideas can be found earlier in the Lasso literature (Roth and Fischer, 2008; Kowalski et al., 2011; Loth, 2011) and also more recently for example in the BLITZ method (Johnson and Guestrin, 2015, 2016) or SDCA variants with (locally) affine losses (Vainsencher et al., 2015). In parallel of these WS approaches where a BCD solver is run many times, first on a small subproblem then on growing ones, it has been proposed to employ so called *safe rules* (El Ghaoui et al., 2012). While a WS algorithm starts a BCD solver using a subset of features, eventually ignoring good ones that shall be later considered, safe rules discard (once and for all) from the full problem some features that are guaranteed to be inactive at convergence.

A number of variants of *safe rules* have been proposed since their introduction, including for SVM-type problems (Ogawa et al., 2013) and we refer to Xiang et al. (2014) for a concise introduction. The most recent versions, called Gap Safe rules, have been applied to a wide range of Lasso-type problems (Fercoq et al., 2015; Ndiaye et al., 2015, 2016). Such rules have the unique property of being convergent, meaning that at convergence only features that map to saturated (dual) constraints remain.

The main contributions of this paper are 1) the introduction of a WS strategy based on an aggressive use of Gap Safe rules which is hence adapted to various Lasso-type estimators, and 2) the demonstration that Gauss-Southwell rules combined with precomputation of Gram matrices can be competitive for the (small) subproblems when looking at running time, and not just in terms of (block) coordinate updates/epochs as previously done in the literature (Nutini et al., 2015; Shi et al., 2016).

The paper is organized as follows: in Section 2, we present how Gap Safe rules can lead to a WS strategy. We then explain how the Gauss-Southwell rule can be employed to reduce computations. Section 4 presents numerical experiments on simulations for GS based inner-solvers, and report time improvements compared to the present state-of-the-art on real datasets.

**Model and notation**

We denote by $[d]$ the set $\{1,\ldots,d\}$ for any integer $d \in \mathbb{N}$. For any vector $u \in \mathbb{R}^d$ and $\mathcal{C} \subset [d]$, $(u)_\mathcal{C}$ is the vector composed of elements of $u$ whose index lies in $\mathcal{C}$, and $\bar{\mathcal{C}}$ is the complementary set of $\mathcal{C}$ in $[d]$. We denote by $\mathcal{S}_B^r \subset [p]$ the row support of a matrix $B \in \mathbb{R}^{p \times q}$ (*i.e.*, the indices of non-zero rows of B). Let $n$ and $p \in \mathbb{N}$ be respectively the number of observations and features and $X \in \mathbb{R}^{n \times p}$ the design matrix. Let $Y \in \mathbb{R}^{n \times q}$ be the observation matrix, where $q$ stands for the number of tasks or classes considered. The Euclidean (resp. Frobenius) norm on vectors (resp. matrices) is denoted by $\|\cdot\|$ (resp. $\|\cdot\|_F$, and the $i$-th row (resp. $j$-th column) of B by $B_{i,:}$ (resp. $B_{:,j}$). The row-wise separable $\ell_{2,1}$ group-norm of a matrix B is written $\|B\|_{2,1} = \sum_i \|B_{i,:}\|$. Let $\Omega$ denote a generic norm, we write $\Omega_*$ its dual norm; for the $\|\cdot\|_{2,1}$ norm this is the $\ell_\infty/\ell_2$ norm $\|B\|_{2,\infty} = \max_i \|B_{i,:}\|$. We denote by $\|B\|_{2,0}$ the number of non-zero rows of B, *i.e.*, the cardinality of $\mathcal{S}_B^r$.

For simplicity we concentrate on quadratic regression losses yet the proposed methodology would readily apply to smooth losses such as the multinomial logistic loss for classification problems with $q$ classes. The penalized multi-task regression estimator that we consider from now on is defined as a solution of the (primal) problem

$$\hat{B}^{(\lambda)} \in \underset{B \in \mathbb{R}^{p \times q}}{\arg\min} \underbrace{\tfrac{1}{2} \|Y - XB\|_F^2 + \lambda \Omega(B)}_{\mathcal{P}^{(\lambda)}(B)} \quad . \tag{1.1}$$

Here, the non-negative $\lambda$ is the regularization parameter controlling the trade-off between data fitting and regularization. The associated dual problem reads (see for instance Ndiaye et al. (2015))

$$\hat{\Theta}^{(\lambda)} = \underset{\Theta \in \Delta_X}{\arg\max} \underbrace{\tfrac{1}{2} \|Y\|_F^2 - \tfrac{\lambda^2}{2} \left\|\Theta - \tfrac{Y}{\lambda}\right\|_F^2}_{\mathcal{D}^{(\lambda)}(\Theta)} \quad . \tag{1.2}$$

where $\Delta_X = \{\Theta \in \mathbb{R}^{n \times q} : \Omega_*(X^\top \Theta) \leq 1\}$ is the (rescaled) dual feasible set. The duality gap for (1.1) is defined by $\mathcal{G}^{(\lambda)}(B, \Theta) := \mathcal{P}^{(\lambda)}(B) - \mathcal{D}^{(\lambda)}(\Theta)$, for $\Theta \in \Delta_X$. When the dependency on $X$ is needed, we write $\hat{B}^{(X,\lambda)}$ (resp. $\hat{\Theta}^{(X,\lambda)}, \mathcal{P}^{(X,\lambda)}(B), \mathcal{D}^{(X,\lambda)}(\Theta)$ and $\mathcal{G}^{(X,\lambda)}(B, \Theta)$) for $\hat{B}^{(\lambda)}$ (resp. $\hat{\Theta}^{(\lambda)}, \mathcal{P}^{(\lambda)}(B), \mathcal{D}^{(\lambda)}(\Theta)$ and $\mathcal{G}^{(\lambda)}(B, \Theta)$).

## 2 From screening rules to working sets

The idea behind safe screening rules is to be able to safely discard features from the optimization process as it is possible to guarantee that the associated regression coefficients will be zero at convergence. The Gap Safe rules proposed first in Fercoq et al. (2015) and later extended in Ndiaye et al. (2015) for the multi-task regression considered here read as follows. For simplicity of the presentation, we now assume that $\Omega = \|\cdot\|_{2,1}$ (other row-wise separable norms could be handled similarly). For a pair of feasible primal-dual variables B and $\Theta$, it is safe to discard feature $j$ in the optimization problem (1.1) if:

$$\left\| X_{:,j}^\top \Theta \right\| + \left\| X_{:,j} \right\| \sqrt{\tfrac{2}{\lambda^2} \mathcal{G}^{(\lambda)}(B, \Theta)} < 1 \quad , \tag{2.1}$$

or equivalently, it is necessary to consider the feature $j$ iff:

$$d_j(\Theta) := \frac{1 - \left\| X_{:,j}^\top \Theta \right\|}{\left\| X_{:,j} \right\|} \leq \sqrt{\tfrac{2}{\lambda^2} \mathcal{G}^{(\lambda)}(B, \Theta)} \quad . \tag{2.2}$$

In other words, the duality gap value allows to define a threshold that shall be compared to $d_j(\Theta)$ in order to safely discard features, and ultimately accelerate solvers. A natural idea, to further reduce running time by limiting problem sizes, while sacrificing safety, is to use the $d_j$'s to prioritize features. One way to formalize this is to introduce a scalar $r \in [0,1]$ and to limit computation to features such that:

$$d_j(\Theta) \leq r\sqrt{\frac{2}{\lambda^2}\mathcal{G}^{(\lambda)}(B,\Theta)} \ . \tag{2.3}$$

Let us consider now this in an iterative strategy. Starting from an initial value of $B_0$ (*e.g.*, $0 \in \mathbb{R}^{p \times q}$ or an approximate solution obtained for a close $\lambda'$), one can obtain a feasible $\Theta_0 \in \Delta_X$, either by using $0 \in \mathbb{R}^{n \times q}$ or by residual normalization (Ndiaye et al., 2015). Assuming $B_0 = 0$, this normalization boils down to scaling $Y/\lambda$ by a constant $\alpha \in [0,1]$ such that $\Omega_*(\alpha X^\top Y/\lambda) = 1$, *i.e.*, choosing[1] $\alpha = \lambda/\lambda_{\max}$, where we write $\lambda_{\max} = \Omega_*(X^\top Y)$.

Given the primal-dual pair $(B_0, \Theta_0)$ one can compute $d_j$ for all features and select the ones to be added to the working set $\mathcal{W}_1$. Then what we will refer to as an *inner solver* can be started on $\mathcal{W}_1$. The iteration for this procedure is as follows: assuming the inner solver returns a primal dual pair $(\tilde{B}_t, \xi_t) \in \mathbb{R}^{p_t \times q} \times \mathbb{R}^{n \times q}$, where $p_t$ is the size of $\mathcal{W}_t$, one can obtain a pair $(B_t, \xi_t)$ by considering that $(B_t)_{\mathcal{W}_t,:} = \tilde{B}_t$ and $(B_t)_{\bar{\mathcal{W}}_t,:} = 0$.

While $\xi_t$ is dual feasible for the subproblem $\mathcal{D}^{(\lambda, X_{\mathcal{W}_t,:})}$ it is not necessarily feasible for the original problem $\mathcal{D}^{(\lambda,X)}$. To obtain a good candidate for $\Theta_t$, Johnson and Guestrin (2015) proposed to find $\Theta_t$ as a convex combination of $\Theta_{t-1}$ and $\xi_{t-1}$:

$$\begin{cases} \alpha_t = \max\{\alpha \in [0,1] : (1-\alpha)\Theta_{t-1} + \alpha\xi_{t-1} \in \Delta_X\} \\ \Theta_t = (1-\alpha_t)\Theta_{t-1} + \alpha_t\xi_{t-1} \end{cases}$$

If $\Theta_0 = 0$ and $B_0 = 0$, computing $\alpha_t$ is equivalent to the residual normalization approach mentioned earlier. Otherwise, $\alpha_t = \min_{j=1,\ldots,p} \alpha^j$ with $\alpha^j = \max\left\{\alpha' \in [0,1] : \left\|X_{:,j}^\top(\alpha'\xi_{t-1} + (1-\alpha')\Theta_{t-1})\right\| \leq 1\right\}$. The computation of $\alpha^j$ has a closed form solution provided in the supplementary material.

So far, we have omitted to detail the strategy to decide which features shall enter the working set at iteration $t$. A first strategy is to set a parameter $r$ and then consider all features that satisfy (2.3). Yet this strategy does not offer a flexible control of the size of $\mathcal{W}_t$. A second strategy, which we use here, is to limit the number of features that shall enter $\mathcal{W}_t$. Constraining the size of $\mathcal{W}_t$ to be at most twice the size of $\mathcal{S}^r_{B_{t-1}}$, one shall keep in $\mathcal{W}_t$ the blocks with indices in $\mathcal{S}^r_{B_{t-1}}$ and add to it the ones in $\bar{\mathcal{S}}^r_{B_{t-1}}$ with the smallest $d_j(\Theta_t)$. This WS growth strategy avoids the two extreme slow cases: adding either too few or too many variables at a time, resulting in too many or too large subproblems respectively. Most contributions we are aware of either add features one by one (*single principal pivoting*) or need tricky bookkeeping to avoid cyclic behavior (Kim and Park, 2010). Our strategy makes the WS grow quickly, while keeping its size comparable to the one of the solution's support, assumed to be moderate by design. Morevoer, when an approximate solution $B^{(\lambda')}$ is available for a close $\lambda'$, *e.g.*, when computing a Lasso path, $p_0$ (the size of the first WS used, which is a parameter of the algorithm) can be replaced by $\|B^{(\lambda')}\|_{2,0}$.

The iterative working set strategy is summarized in Algorithm 1. When this working set algorithm is combined with the block coordinate descent inner solver described in Section 3, we call it A5G (which stands for AGGressive Gap, Greedy with Gram).

When $q = 1$ and one considers only $\ell_1$ regularized problems the strategy just explained is similar to the Blitz algorithm (Johnson and Guestrin, 2015). Indeed in the $\ell_1$ case, the $d_j$'s boil down to the computation of the distance to the constraints for the dual problem (Johnson and Guestrin, 2015). For the $\ell_{2,1}$ norm considered here the computation of the distance from $\Theta_t$ to the set $\{\Theta \in \mathbb{R}^{n \times q} : \left\|X_{:,j}^\top \Theta\right\| = 1\}$ involves projection on ellipsoids for which no closed-form solution exist[2]. However, viewing BLITZ as an aggressive Gap Safe screening strategy allows for immediate adaptation of (2.3) to more generic sparse penalties for which Gap Safe rules have been derived. We illustrate this here with the multi-task Lasso. Following Ndiaye et al. (2015), for the $\ell_{2,1}$ regularization the quantity $d_j$ reads:

$$d_j(\Theta_t) := \frac{1 - \left\|X_{:,j}^\top \Theta_t\right\|}{\left\|X_{:,j}\right\|} \ . \tag{2.4}$$

---

[1] We only consider the case where $\lambda \leq \lambda_{\max}$, as 0 is a trivial solution of (1.1) otherwise.

Table 1: Computation cost (first line) for one block update using different BCD strategies (cyclic or GS-r greedy rule with or without precomputation of the Gram matrix). Storage cost (second line) for efficient updates. It accounts for the regression coefficients ($pq$), the precomputation of the norms of the regressors ($p$), the eventual precomputation of the Gram matrix ($p^2$) and current value of the gradients ($pq$) (with precomputation of the Gram matrix) or the residual ($nq$) (without precomputation of the Gram matrix).

| BCD strategy | Cyclic | GS-r | Cyclic (Gram) | GS-r (Gram) | GS-rB (Gram) |
|---|---|---|---|---|---|
| Computation | $nq$ or $2nq$ | $npq$ or $npq + nq$ | $q$ or $q + pq$ | $2pq$ | $(p + B)q$ |
| Storage | $nq + p + pq$ | $nq + p + pq$ | $2pq + p + p^2$ | $2pq + p + p^2$ | $2pq + p + p^2$ |

Note that our focus on multi-task Lasso is due to a particular concern on M/EEG applications (see Section 4), but our approach is well-defined as soon as Gap Safe screening rules exists, *e.g.*, for Sparse-group Lasso or $\ell_1$ regularized logistic regression. This flexibility is an asset compared to other strategies such as the one proposed by Kim and Park (2010) which rewrite the Lasso as a quadratic programming problem.

Another important aspect of our algorithm is that we do not require the exact solution of the subproblems to be computed: we solve the subproblem restricted to $\mathcal{W}_t$ up to a duality gap equal to a fraction $\epsilon \in [0,1]$ of the current duality gap of the whole problem. We address the choice of this parameter, as well as the global stopping criterion $\bar{\epsilon}$, in Section 3.4.

Now that we have detailed the WS strategy we perform, we are ready to address the choice of the inner solver that minimizes (1.1) restricting $X$ to the features in the set $\mathcal{W}_t$.

## 3 Block Coordinate Descent (BCD) as inner solver

In our context (with $\Omega = \|\cdot\|_{2,1}$) the function we aim at optimizing has the following form: $\mathcal{P}^{(\lambda)}(B) = f(B) + \lambda \sum_{j=1}^p \|B_j\|$, where $f(\cdot) = \|Y - X \cdot\|_F^2 / 2$. In this section we simply write $B_j \in \mathbb{R}^{1 \times q}$ to refer to the row $B_{j,:}$. When considering a block coordinate descent algorithm, one sequentially updates at step $k$, a single block (here row) $j_k$ of B. The strategy to choose $j_k$ is discussed in Section 3.1. For our problem, the block update rule proceeds as follows:

$$B_{j_k}^k = \mathcal{T}_{j_k, L_{j_k}}(B^{k-1}) \ , \tag{3.1}$$

---
[2] Note that for general norms, such projection would become even more intricate

**Algorithm 1** A5G

**input** : $X, Y, \lambda$
**param**: $p_0 = 100, \xi_0 = Y/\lambda, \Theta_0 = 0_{n,q}, B_0 = 0_{p,q},$
$\bar{\epsilon} = 10^{-6}, \underline{\epsilon} = 0.3$

**for** $t = 1, \ldots, T$ **do**
    $\alpha_t = \max\{\alpha \in [0,1] : (1-\alpha)\Theta_{t-1} + \alpha\xi_{t-1} \in \Delta_X\}$
    $\Theta_t = (1-\alpha_t)\Theta_{t-1} + \alpha_t\xi_{t-1}$
    $g_t = \mathcal{G}^{(X,\lambda)}(B_{t-1}, \Theta_t)$      // global gap
    **if** $g_t \leq \bar{\epsilon}$ **then**
        | Break
    **for** $j = 1, \ldots, p$ **do**
        Compute $d_j^t = (1 - \|X_{:,j}^\top \Theta_t\|)/\|X_{:,j}\|$
        // safe screening:
        Remove $j^{th}$ column of $X$ if $d_j^t > \sqrt{2g_t/\lambda^2}$
    Set $(d^t)_{\mathcal{S}_{B_{t-1}}^r} = -1$    // keep active features
    $p_t = \max(p_0, \min(2\|B_{t-1}\|_{2,0}, p))$    // clipping
    $\mathcal{W}_t = \{j \in [p] : d_j^t \text{ among } p_t \text{ smallest values of } d^t\}$
    // Approximately solve sub-problem :
    Get $\tilde{B}_t, \xi_t \in \mathbb{R}^{p_t \times q} \times \Delta_{X_{:,\mathcal{W}_t}}$ s.t. $\mathcal{G}^{(X_{:,\mathcal{W}_t},\lambda)}(\tilde{B}_t, \xi_t) \leq \underline{\epsilon} g_t$
    Set $B_t \in \mathbb{R}^{p \times q}$ s.t. $(B_t)_{\mathcal{W}_t,:} = \tilde{B}_t$ and $(B_t)_{\bar{\mathcal{W}}_t,:} = 0$.

**return** $B_t$

---

where for all $j \in [p]$ the partial gradient over the $j^{th}$ block $\nabla_j f$ is assumed to be $L_j$-Lipschitz (where $L_j = \|X_{:,j}\|^2$ is a possible choice),

$$\mathcal{T}_{j,L}(B) := \text{prox}_{\frac{\lambda}{L}\|\cdot\|}\left(B_j - \frac{1}{L}\nabla_j f(B)\right), \quad (3.2)$$

with for any $z \in \mathbb{R}^q$ and $\mu > 0$,

$$\text{prox}_{\mu\|\cdot\|}(z) = \underset{x \in \mathbb{R}^q}{\arg\min} \frac{1}{2}\|z - x\|^2 + \mu\|x\|. \quad (3.3)$$

For multi-task problems the proximal computation is simply a block soft-thresholding step, see *e.g.*, Parikh et al. (2013, p. 65):

$$\text{prox}_{\mu\|\cdot\|}(z) = \text{BST}(z, \mu) := \left(1 - \frac{\mu}{\|z\|}\right)_+ z. \quad (3.4)$$

where for any real number $a$, $(a)_+ = \max(0, a)$ refers to the positive part of $a$.

## 3.1 Visiting strategies

In this section we present several visiting strategies, *i.e.*, strategies on how to pick the $j_k^{th}$ block to be updated in Equation (3.1). The first two standard methods rely on update strategies that are fixed prior to any computation, and the following ones are Gauss-Southwell variants.

**Cyclic strategies** The simplest and most common strategy consists in selecting the coordinate in a cyclic manner:

$$\text{Pick} \quad j_k = (k \mod p) + 1 \ . \tag{3.5}$$

This rule can be easily modified by permuting the visiting order of the blocks after each epoch[3], see Beck and Tetruashvili (2013); Beck et al. (2015) for a theoretical analysis.

**Randomized strategies** Another popular strategy is to draw $j_k$ according to a fixed probability distribution over the blocks. Here, we only mention uniform random selection, since in common statistical applications the dataset is normalized prior to any computation, so that $\|X_{:,j}\| = \sqrt{n}$ for all $j \in [p]$ (making most variants such as the one proposed by Nesterov (2012) identical in our case):

$$\text{Pick} \quad j_k = j \quad \text{with probability} \quad \frac{1}{p}, \quad \forall j \in [p] \ . \tag{3.6}$$

In our practical experiments, this strategy has always been outperformed by the cyclic strategy when looking at running time. For this reason we do not present comparisons for randomized variants in the experiment Section 4. Note that similar conclusions are reached by Shi et al. (2016) in terms of number of epochs.

## 3.2 Greedy / Gauss-Southwell strategies

Here we follow the presentation and terminology introduced by Tseng and Yun (2009); Nutini et al. (2015); Shi et al. (2016), and we remind some of the classical variants of Gauss-Southwell (GS), *i.e.*, greedy block coordinate descent strategies. Contrary to strategies presented in Section 3.1, these variants aim at dynamically identifying the "best" block to be updated. They differ only in the "optimal" criterion considered for assessing this choice. Note that the greedy variants might be interesting in the context of sparse regularization as they focus on updating blocks that correspond to the targeted support, which is often expected to be very small. This property will be illustrated for instance by Figure 1 and discussed in the experiments section.

**GS-MBI** The Maximum Block Improvement (GS-MBI) strategy was analyzed in Chen et al. (2012); Li et al. (2015) and consists in picking the block making the primal objective function $\mathcal{P}^{(\lambda)}$ decrease the most when fixing the other blocks. This means that one picks $j_k$ according to

$$\text{Pick} \ j_k \in \operatorname*{arg\,min}_{j \in [p]} \min_{\mathrm{B}_j \in \mathbb{R}^q} \mathcal{P}^{(\lambda)}(\mathrm{B}_1^{k-1}, \ldots, \mathrm{B}_j, \ldots, \mathrm{B}_p^{k-1}) \ . \tag{3.7}$$

This rule is usually costly, since it requires recomputing the primal objective function, in particular when the residual matrix is not stored (see Section 3.4 where a detailed discussion is given).

**GS-r** This rule chooses the block that maximizes the length of the step performed, meaning the block considered is the one satisfying

$$\text{Pick} \quad j_k \in \operatorname*{arg\,max}_{j \in [p]} \max_{\mathrm{B}_j \in \mathbb{R}^q} \left\| \mathcal{T}_{j,L_j}(\mathrm{B}^{k-1}) - \mathrm{B}_j^{k-1} \right\| \ . \tag{3.8}$$

---

[3]where an epoch refers to a pass over all $p$ blocks

With $\mathcal{Q}_{j,B}(z) = z^\top \nabla_j f(B) + \frac{L_j}{2}\|z\|^2 + \lambda \|B_j + z\|$, this strategy is equivalent to picking the block that maximizes the length of the step performed by an update obtained substituting a quadratic surrogate to $f$:

$$\text{Pick} \quad j_k \in \arg\max_{j \in [p]} \left\| \arg\min_{z \in \mathbb{R}^q} \mathcal{Q}_{j,B^{k-1}}(z) \right\| . \qquad (3.9)$$

**GS-q** This rule is similar to the previous one, but picks the coordinate that maximizes the improvement of the surrogate function for the objective:

$$\text{Pick} \quad j_k \in \arg\min_{j \in [p]} \left( \min_{z \in \mathbb{R}^q} \mathcal{Q}_{j,B^{k-1}}(z) \right) . \qquad (3.10)$$

Note that for a data-fitting term $f(B) = \|Y - XB\|_F^2/2$ the two strategies GS-q and GS-r are the same.

**GS-rB** In our experiments on Lasso-type problems, GS-r was usually the rule performing most efficiently. Building on this fact, we have considered a variant that can further reduce its computational cost by performing the picking only over small batches. We call GS-rB such a variant strategy of GS-r. Here, instead of aiming at finding the best block (in terms of largest update move) among the $p$ possible ones, we regroup the blocks in contiguous batches of size $B$. This allows scanning only $B$ features instead of $p$ for each update. In practice we chose to use deterministic batches, where the $k^{th}$ batch contains the blocks of indices between $kB$ and $(k+1)B$. Experiments were done with $B = 10$ as it provided already satisfying speed-ups over the pure GS-r strategy (see supplementary material for more insights on the influence of the batch size $B$ on the convergence speed).

As it excludes some features of the subproblems, it is not obvious that Algorithm 1 converges. It is however the case when used with the GS-r rule in the inner solver, as we show in the supplementary material.

### 3.3 Gram matrix precomputation

As the best block is selected for each update, greedy strategies decrease the number of epochs needed to reach a targeted accuracy. Yet, this comes at the price of additional computation for selecting the best block. When the latter step is costly, the theoretical benefits of GS rules is not necessarily reflected in (real) computing times. In fact, for naive implementations, GS rules show no benefit in the resolution for sparse regression. This is illustrated for instance in Figure 2.

Diving more into the block updating rule, a single block coordinate descent update consists in applying Equation (3.1) with the choice $L_j = \|X_{:,j}\|^2$ and

$$\mathcal{T}_{j,L_j}(B) = \text{BST}\left( B_j - \frac{1}{L_j} X_{:,j}^\top (XB - Y), \frac{\lambda}{L_j} \right) , \qquad (3.11)$$

where BST is the block soft-thresholding operator reminded in (3.4). For efficiency (when the Gram matrix $Q = X^\top X$ is not precomputed), the residual $R^k = Y - XB^k$ is maintained, leading to the

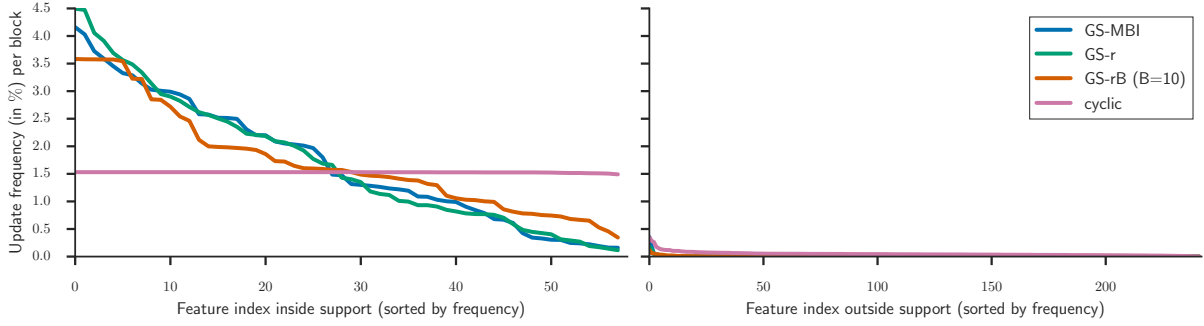

Figure 1: Evaluation of the proportion of updates by coordinate, for a Lasso case, with Gaussian/Toeplitz design simulated dataset ($n = 1000, p = 1000$). The proportion is displayed for blocks inside (left) and outside (right) the support of the converged solution (its support size being 58), ranking blocks by decreasing frequency for each method. All methods are with precomputed Gram matrix.

update:
$$\begin{cases} R & \leftarrow R^{k-1} + X_{:,j} B_j^{k-1} \quad \text{if} \quad B_j^{k-1} \neq 0 \\ B_j^k & \leftarrow \text{BST}\left(\frac{1}{L_j} X_{:,j}^\top R, \frac{\lambda}{L_j}\right) \\ R^k & \leftarrow R - X_{:,j} B_j^k \quad \text{if} \quad B_j^k \neq 0 \end{cases}. \quad (3.12)$$

The main costs of this update are a product between a size $n$ vector and a size $n \times q$ matrix and 0, 1 or 2 rank one updates of $n \times q$ matrices.

In a context where one can precompute and store the Gram matrix $Q = X^\top X = [Q_1, \ldots, Q_p]$, an adaptation of the update can provide huge benefits. It consists in maintaining the gradients $H^k = X^\top(XB^k - Y) \in \mathbb{R}^{p \times q}$, rather than the residuals. The BCD update can then be written:

$$\begin{cases} \delta B_j & \leftarrow \text{BST}\left(B_j^{k-1} - \frac{1}{L_j} H_j^{k-1}, \frac{\lambda}{L_j}\right) - B_j^{k-1} \\ B_j^k & \leftarrow B_j^{k-1} + \delta B_j \quad \text{if} \quad \delta B_j \neq 0 \\ H^k & \leftarrow H^{k-1} + Q_j \delta B_j \quad \text{if} \quad \delta B_j \neq 0 \end{cases}. \quad (3.13)$$

If the update is not 0, the main cost is a rank one update of a $p \times q$ matrix (the third line). Otherwise, it is the computation of the norm of a size $q$ vector : an interesting aspect of this conditional update is that the cost of visiting a block is small when the block does not change after the update. This is one of the key reasons for the particularly good performance of the cyclic rule for sparse regularization: updates are cheaper when coefficients are zeros. This is illustrated in Figure 1 where the number of gradient updates by the cyclic rule over the support is particularly small: even if the blocks are visited, the gradient update is rarely performed.

Moreover, with a precomputation of the Gram matrix $Q = X^\top X$, the update rules are efficiently performed, and it become beneficial to apply GS strategies for practical speed-ups.

The computation cost for one block update using different selection strategies is summarized in Table 1. This table also contains details on the memory allocations necessary to handle the blocks, the (block-wise) Lipschitz constants and the maintained residual or gradients (depending on whether a precomputation of $Q$ is performed).

## 3.4 Stopping criterion

The stopping criterion is a key element when comparing implementations, especially when trying to be fair between various methods. Providing an efficient one is therefore of high practical interest. Thanks to strong duality, *cf.* for instance Bauschke and Combettes (2011), the duality gap provides a natural stopping criterion. Indeed, for any primal dual pair $B^k \in \mathbb{R}^{p \times q}$ and $\Theta^k \in \Delta_X$, the following property holds: if $\mathcal{G}^{(\lambda)}(B^k, \Theta^k) = \mathcal{P}^{(\lambda)}(B^k) - \mathcal{D}^{(\lambda)}(\Theta^k) \leq \epsilon$ then $\mathcal{P}^{(\lambda)}(B^k) - \mathcal{P}^{(\lambda)}(\hat{B}) \leq \epsilon$. Thus, stopping an algorithm when $\mathcal{G}^{(\lambda)}(B^k, \Theta^k) \leq \epsilon$ provides an $\epsilon$-solution for the primal problem. A standard strategy is to compute the gap every $f^{ce}$ block updates, for example $f^{ce} = 10p$ corresponds to every 10 epochs on the data for a cyclic rule. However, this means computing the duality gap less frequently for methods with a larger cost per update (*e.g.*, for the GS strategies), leading to solvers that have converged, yet continue running.

Yet, the solution is not to evaluate the duality gap blindly more often as it can be costly. When using updates with precomputed Gram matrix, the residuals $Y - XB^k$ are not stored (only the gradient is stored then). There is therefore a cost in evaluating the primal function.

Moreover, a (dual) feasible point is required to evaluate the dual objective. We obtain it by residual normalization, *i.e.*,

$$\Theta^k = \frac{R^k}{\max(\lambda, \|X^\top R^k\|_{2,\infty})} \ . \tag{3.14}$$

Overall, a duality gap evaluation requires $n \cdot q \cdot \|B\|_{2,0}$ operations with a precomputed Gram matrix, which corresponds for instance to the cost of $n$ updates with the cyclic strategy.

To avoid such a cost, a possible remedy is to monitor the primal decrease, and trigger a duality gap computation only when this might be beneficial. Let us write $\delta^{k+1} := \mathcal{P}^{(\lambda)}(B^k) - \mathcal{P}^{(\lambda)}(B^{k+1})$ the primal decrease at step $k+1$. Assuming $\delta^{k+1} > \epsilon$ (*i.e.*, a decrease of the primal was at least of amplitude $\delta^{k+1}$), then $\mathcal{P}^{(\lambda)}(B^k) > \mathcal{P}^{(\lambda)}(\hat{B}) + \delta$, so the $k^{th}$ iterate was not $\epsilon$-optimal, and there was no need to compute the duality gap for such a point. In particular, this is true if $\lambda(\|B^k_{j_k}\| - \|B^{k+1}_{j_k}\|) > \epsilon$. A heuristic variant could be simply to prevent computing the duality gap when $\lambda(\|B^k_{j_k} - B^{k+1}_{j_k}\|) > \epsilon)$. Such a trick is used in the default Lasso and multi-task Lasso solvers in scikit-learn (Pedregosa et al., 2011).

## 4 Experiments

In this section we present some experimental results both on simulations using poorly conditioned Toeplitz designs (see for instance Bühlmann and van de Geer (2011, p. 23) for more details) as well as on two real-world datasets. Results on real data are shown for the Lasso ($q = 1$) and multi-task Lasso estimators. The implementation was done in Python, using Cython/C for block coordinate descent inner solvers.

### 4.1 Experiments on BCD inner solvers

First, we evaluate empirically the behavior of the most popular GS strategies on the Lasso case ($q = 1$), in particular when the Gram matrix can be precomputed.

We run several BCD solvers with different selection rules (cyclic, GS-MBI, GS-r and batch GS-rB with $B = 10$) on a $n = 300, p = 300$ strongly-correlated Toeplitz simulated design ($\rho = 0.99$), until a duality gap of $10^{-6}$ is reached. This design is supposed to model a subproblem, hence its moderate size; experiments on larger Toeplitz designs are presented in the supplementary. During the optimization process, we keep track of the number of times each coefficient value has changed. Figure 1 shows that although the cyclic strategy visits all block with equal frequencies ($1/p$), it seldom changes coefficients outside of the converged solution support: this reflects the fact that the computing time, even for the cyclic rule, is spent mostly on "useful" updates when the conditional update in (3.13) is used with a precomputed Gram matrix. This avoids useless gradient or residual updates when the coefficients are untouched, typically when they remain equal to zero. On the contrary, as GS strategies pick the block whose update is the "best" (block-wise for GS-r or objective-wise for GS-MBI), the selected block is always updated, unless the solver has already reached convergence. We also see that GS-r and GS-rB have similar distributions of updates, illustrating that exploring only a fraction of the blocks is an efficient proxy for the GS-r strategy. Additional experiments have shown that among the variants of GS strategies, GS-r is a satisfactory choice in practice for Lasso-type problems.

To demonstrate the computational benefits of greedy strategies when the Gram matrix is precomputed, we run several Lasso solvers on the same dataset evaluating the duality gap periodically. Figure 2 shows that for such a setting with a moderate number of features, the use of the Gram matrix makes the greedy selection rule competitive with the cyclic rules[4].

Precomputing the Gram matrix consistently helps (darker lines lower than the others), and the gain is even more striking for the GS-r rule. Indeed, maintaining the vector of gradients at each update makes the cost of the GS-r and GS-rB rule tractable, and the large coefficient updates induced compensate for their cost. This observation is particularly relevant for early iteration as the gains are the strongest during this stage. Below a dual gap of $10^{-4}$, one can observe the computational benefits of the GS-rB rule that makes the cost of one update decrease from $2p$ to $p + B$ floating point operations (FLOPS) (*cf*. Table 1).

What can also be observed here is that the cyclic rule with Gram matrix precomputation is rather competitive. Looking at Table 1 this can be explained by the fact that the cost of one visit is either 1 or $1 + p$ (assuming $q = 1$). Figure 2 suggests that many visits have a low computational cost (*i.e.*, 1), which is conform with Figure 1, showing the frequencies of updates. Again the cyclic strategy rarely triggers an update of the gradient for a feature outside of the (converged) active set. The cost of one block visit (which includes both selection and update, as well as the eventual gradeints or residuals update) for the different strategies, with or without Gram matrix precomputation, are summarized in Table 1.

## 4.2 Solvers with working sets

We now evaluate the performance of the WS strategy detailed above on two datasets. First we consider the Lasso problem which allows us to compare our implementation to the state-of-the-art C++ implementation of BLITZ by Johnson and Guestrin (2015)[5]. We only compare to BLITZ, since extensive experiments in Johnson and Guestrin (2015) showed that it is currently the fastest solver for the Lasso. Note that BLITZ implementation uses a heuristic parameter called tolerance as an additional stopping criterion, and rescale the duality gap by the current dual objective. We use the default value for this parameter ($10^{-4}$). In all our experiments we set $\bar{\epsilon} = 10^{-6}$, which means that the solver stops as soon as the duality gap goes below $10^{-6}$ on the full problem. We set $\underline{\epsilon} = 0.3$ so that on the inner subproblems, the solvers have to reach a duality gap which is 30% lower than the current duality gap estimate.

---

[4]We have mostly focused on middle sized problems, since with WS strategies those are relevant sizes for the sub-problems solved, *cf*. supplementary material for an example on the Leukemia dataset

[5]https://pypi.python.org/pypi/blitzl1/

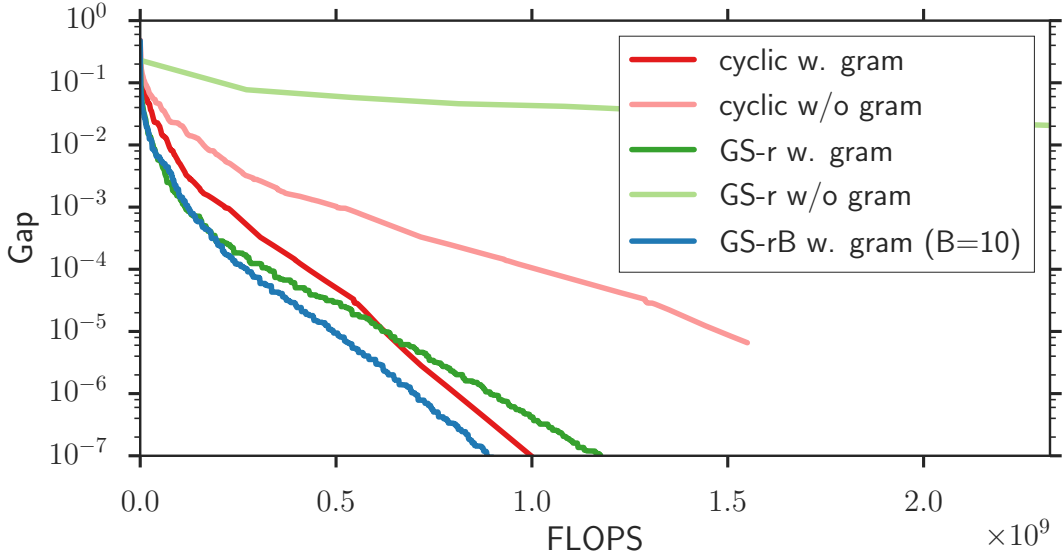

Figure 2: Duality gap as a function of the number of floating point operations (FLOPS) counted as in Table 1. Design ($n = 300$, $p = 300$) is a Toeplitz matrix with strong correlation ($\rho = 0.99$) with a low regularization parameter. One can see that Gram matrix precomputation helps significantly and that greedy strategies are competitive in terms of computation.

Figure 3 presents the duality gap as a function of time on the standard Leukemia dataset. We can observe that our implementation reaches comparable performance with the BLITZ C++ implementation, which is itself significantly better than the scikit-learn implementation (Pedregosa et al., 2011) that does not make use of any working set. Experiments for other values of $\lambda$ are included in the supplementary and show the same performance.

Figure 4 presents results on a multi-task Lasso problem, relevant for brain imaging with electroencephalography (EEG) and magnetoencephalography (MEG). EEG and MEG are brain imaging modalities that allow to localize active regions in the brain. The $Y$ observations are multivariate time-series and the design matrix is obtained from the physics of the problem. Here $n$ corresponds to the number of sensors, $q$ corresponds to the number of time instants and $p$ to the number of locations in the brain. The multi-task Lasso estimator allows here to identify brain activity which is stable on a short time interval Ou et al. (2009). In this experiment, we use event related field data (from the MNE dataset, see Gramfort et al. (2014)) following an auditory stimulation in the left ear, in fixed orientation setting. There are 302 MEG sensors, $p = 7498$ candidate brain locations and $q = 181$ which corresponds to a time interval of about 300 ms. We set $\lambda = 0.1\lambda_{\max}$, which leads to 24 blocks with non-zero coefficients at convergence, meaning 24 brain locations active in the model.

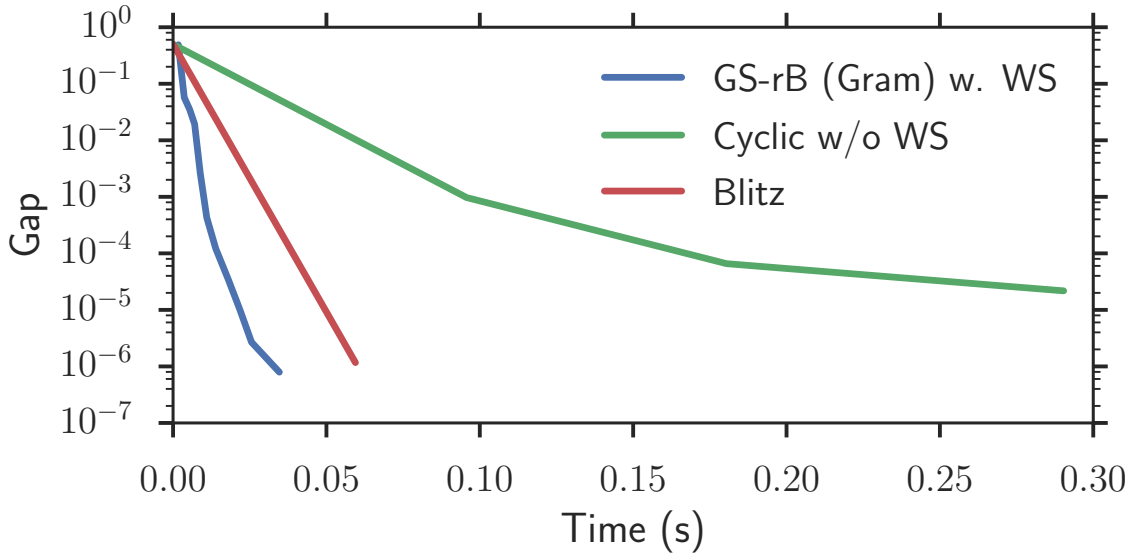

Figure 3: Duality gap as a function of time for the Lasso on the standard Leukemia dataset ($n = 72, p = 7129$) using $\lambda = 0.01\|X^\top Y\|_{2,\infty}$. Methods compared are the cyclic BCD from scikit-learn (Cyclic w/o WS), the C++ implementation of Blitz as well as our WS approach combined with the GS-rB rule ($B = 10$) with precomputation of the Gram matrix. Both WS approaches outperform the plain BCD solver.

## 5 Conclusion and future work

In this paper we have proposed a connection between Gap Safe screening rules and working set (WS) strategies, such as Blitz, in particular to tackle more generic learning problems under sparsity assumptions such as the multi-task regression with $\ell_{2,1}$ regularization. We have provided a thorough analysis of block coordinate descent variants, when used as inner solver for WS methods. We have illustrated the benefit of precomputing the Gram matrix in such a context. Precomputations allow Gauss-Southwell (GS) variants to reach comparable performance to cyclic updates, not only in terms of epochs but also in terms of computing time. To our knowledge, our implementation is the first to demonstrate timing performance for GS rules. In particular, a GS variant we coined GS-rB, relying on restricting the search of the best update cyclically over small batches of blocks has provided the best compromise. Last but not least the conjunction of WS strategies with GS methods reached noticeable speed-ups with respect to standard open source implementation available. Among possible improvements we believe that more refined batch GS strategies could provide improved performance, possibly by performing several pass over the same batches. Additionally, improving the efficiency of the stopping criterion strategies would be another venue for future research, either for the inner solver ($\underline{\epsilon}$), or for the WS method ($\overline{\epsilon}$). Finally, for simplicity we have considered at most doubling the size of the WS, but the impact of this growing factor might need more insights.

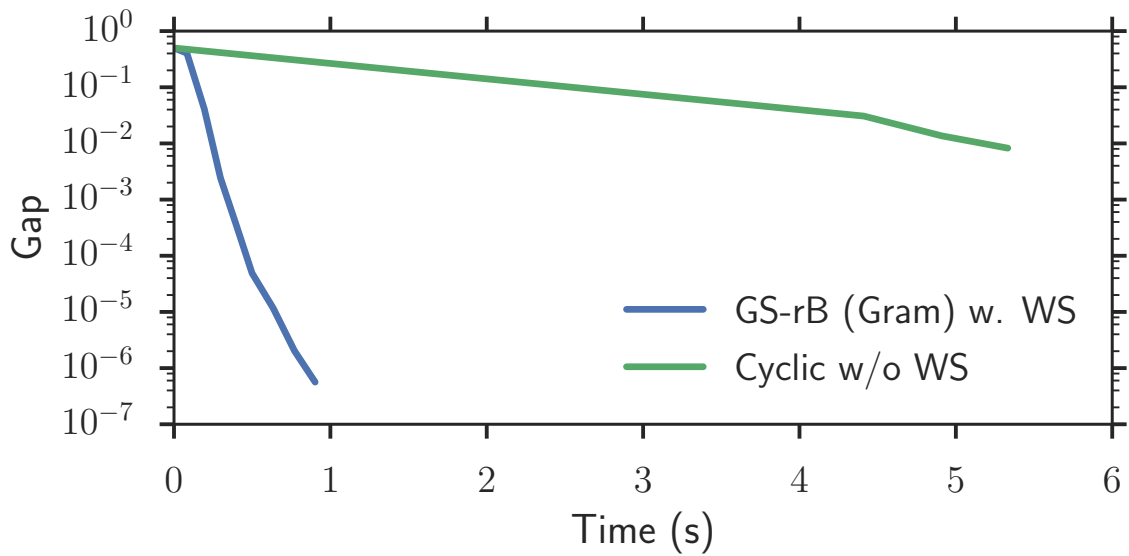

Figure 4: Duality gap as a function of time for the multi-task Lasso on MEG data ($n = 302, p = 7498, q = 181$) using $\lambda = 0.1 \|X^\top Y\|_{2,\infty}$. The cyclic BCD from `scikit-learn` is compared to the WS approach combined with the GS-rB rule ($B = 10$) with precomputation of the Gram matrix. The proposed WS approach outperforms the plain BCD solver.

## Acknowledgments

This work was funded by the ERC Starting Grant SLAB ERC-YStG-676943.

## A Proof of convergence

We show that Algorithm 1 converges when used with the GS-r rule and the predictors are normalized, using the fact that the GS-r applied to the whole problem converges Tseng and Yun (2009). At iteration $t$, let $j$ be the (potentially non unique) optimal variable in the sense of the GS-r criterion. If $j \in \mathcal{S}^r_{B_{t-1}}$, i.e., $(B_t)_j \neq 0$, then by construction $j$ is included in $\mathcal{W}_t$. On the contrary, if $j \notin \mathcal{S}^r_{B_{t-1}}$, since it is optimal in the GS-r sense, for all $j' \notin \mathcal{S}^r_{B_{t-1}}$,

$$\left\| \text{BST}(X^\top_{:,j}(Y - XB_{t-1}), \lambda) \right\| \geq \left\| \text{BST}(X^\top_{:,j'}(Y - XB_{t-1}), \lambda) \right\|, \tag{A.1}$$

or equivalently,

$$\left\| X^\top_{:,j}(Y - XB_{t-1}) \right\| \geq \left\| X^\top_{:,j'}(Y - XB_{t-1}) \right\|. \tag{A.2}$$

So if $\Theta_t$ is chosen to be the residuals rescaled as in Ndiaye et al. (2015), we have $d_j(\Theta_t) \leq d_{j'}(\Theta_t)$, and the feature $j$, having the lowest $d_j(\Theta_t)$ amongst $\mathcal{S}^r_{B_{t-1}}$, is included in $\mathcal{W}_t$. In the case where $j$ is not unique, at least one GS-r optimal feature is included in $\mathcal{W}_t$.

In both cases $j \in \mathcal{W}_t$, and since it is the optimal GS-r variable for the whole problem, it is also the optimal variable for the subproblem restricted to $\mathcal{W}_t$, hence it is the first one updated in the inner GS-r loop. Hence Algorithm 1 performs the GS-r optimal update at the beginning of each subproblem, ensuring the convergence following Tseng and Yun (2009).

## B Implementation details

### B.1 Computation of $\alpha_t$ in Algorithm 1

We remind that $\alpha_t = \max\{\alpha \in [0,1] : (1-\alpha)\Theta_{t-1} + \alpha\xi_{t-1} \in \Delta_X\}$.

In particular since $\Delta_X = \left\{ \Theta \in \mathbb{R}^{n \times q} : \forall j \in [p] \left\| X^\top_{:,j}\Theta \right\| \leq 1 \right\}$, then we have that $\alpha_t = \min_{j=1,\ldots,p} \alpha^j$, with

$$\alpha^j = \max\left\{ \alpha' \in [0,1] : \left\| X^\top_{:,j}(\alpha'\xi_{t-1} + (1-\alpha')\Theta_{t-1}) \right\| \leq 1 \right\}.$$

If constraint $j$ is satisfied by $\xi_{t-1}$, since $\theta_{t-1}$ satisfies all constraints by construction, we have that $\alpha^j = 1$. We can thus reduce the computation of $\alpha^j$ to constraints violated by $\xi_{t-1}$, for which it requires solving the following quadratic equation in $\alpha$:

$$\left\| \alpha X^\top_{:,j}\xi_{t-1} + (1-\alpha)X^\top_{:,j}\Theta_{t-1} \right\|^2 = 1$$

$$\Leftrightarrow \alpha^2 \left\| X^\top_{:,j}(\xi_{t-1} - \Theta_{t-1}) \right\|^2 + 2\alpha(X^\top_{:,j}(\xi_{t-1} - \Theta_{t-1}))(X^\top_{:,j}\Theta_{t-1})^\top + \left\| X^\top_{:,j}\Theta_{t-1} \right\|^2 - 1 = 0.$$

Hence, with the discriminant being

$$\text{disc} = 4\left( \left(X^\top_{:,j}(\xi_{t-1} - \Theta_{t-1})(X^\top_{:,j}\Theta_{t-1})^\top\right)^2 + \left(1 - \left\| X^\top_{:,j}\Theta_{t-1} \right\|^2\right)\left\| X^\top_{:,j}(\xi_{t-1} - \Theta_{t-1}) \right\|^2 \right),$$

then

$$\alpha^j = \begin{cases} 1, & \text{if } \|X_{:,j}^\top \xi_{t-1}\| \leq 1, \\ \frac{-2(X_{:,j}^\top(\xi_{t-1}-\Theta_{t-1}))(X_{:,j}^\top\Theta_{t-1})^\top + \sqrt{\text{disc}}}{2\|X_{:,j}^\top(\xi_{t-1}-\Theta_{t-1})\|^2}, & \text{otherwise} \end{cases}. \quad (B.1)$$

In the Lasso case ($q = 1$), this boils down to:

$$\alpha^j = \begin{cases} 1, & \text{if } |X_{:,j}^\top \xi_{t-1}| \leq 1 \\ \frac{1-|X_{:,j}^\top\Theta_{t-1}|}{|X_{:,j}^\top(\xi_{t-1}-\Theta_{t-1})|}, & \text{otherwise} \end{cases}. \quad (B.2)$$

## C Additional experiments

### C.1 Influence of batch size

Figure 5 illustrates on synthetic data the influence of the batch size $B$ in the GS-rB method. The simulation scenario is a Toeplitz design with similar noise level ($\sigma = 10$) as previously discussed, with low ($\lambda_{\max}/500$) and high ($\lambda_{\max}/20$) regularization parameters. In this setting, batch sizes around 10 % of $p$ (e.g., $B = 100$ or 300) achieve the most interesting rate. So we use $B = 10$ when we use a default value of 100 for $p0$ (the minimal size of the WS). A batch size depending on the WS size could be interesting, but is not investigated here;

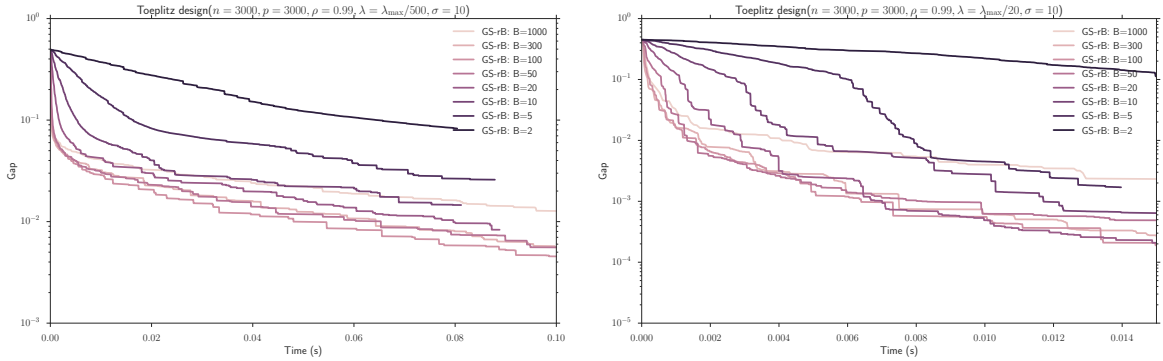

Figure 5: Lasso problem: duality gap as a function of time for different batch sizes in greedy strategies (GS-rB) in CD method. The design matrix ($n = 3000$, $p = 3000$) is Toeplitz with high correlation ($\rho = 0.99$) and a low regularization parameter.

### C.2 Working sets sizes

Figure 6 illustrates that the working sets constructed in the outer loop of our WS remain of moderate size. Indeed for the Leukemia dataset ($n = 72, p = 7129$), even with a very small regularization term $\lambda = \lambda_{\max}/100$, the sizes of the subproblems encountered before convergence ($\bar{\epsilon} = 10^{-6}$) remain moderate (smaller than 200 in all cases). This illustrates the fact that computation and storage of the Gram matrix are possible for the inner problems. We also see that the WS constructed are much smaller than the *sure active sets*, which consist of the variables that cannot be discarded by safe screening rules.

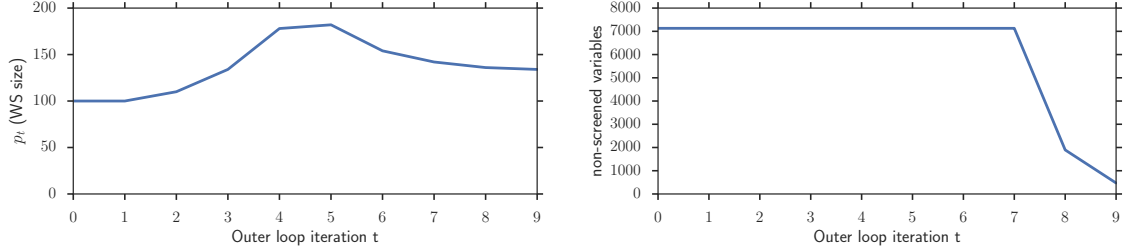

Figure 6: Left: Number of features included in the working set at each outer loop iteration on the Leukemia dataset, for a Lasso problem with $\lambda = \lambda_{\max}/100$. Right: number of non screened variable when dynamic safe screening rules are used, for the same problem.

### Comparison with Blitz for various $\lambda$

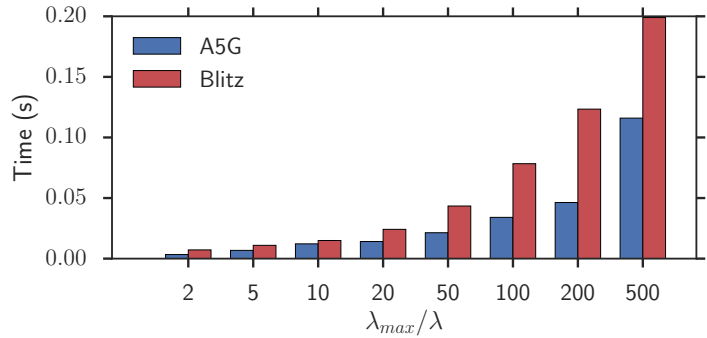

Figure 7: Time to reach a duality gap of $10^{-6}$ as a function of $\lambda$.

Figure 7 shows that the results presented in Figure 3 for the Leukemia dataset also hold for a wide range of values of $\lambda$, chosen in an approximately logarithmic fashion between $\lambda_{max}/2$ and $\lambda_{max}/500$.

### Influence of $\underline{\epsilon}$

Figure 8 shows the influence of the parameter $\underline{\epsilon}$. We remind that the subproblems are solved up to a duality gap of $\underline{\epsilon}\mathcal{G}^{\lambda,X}(\mathrm{B}_{t-1},\Theta_{t-1})$. Since it is too costly to compute the duality gap after each update, a gap computation frequency must be used. As in Ndiaye et al. (2015), we compute the duality gap every $10p_t$ updates, which means every 10 passes over all active blocks in the cyclic update rule. This cancels the impact that an $\underline{\epsilon}$ too close to 1 would have, since it imposes a minimal number of updates.

On the other hand, we also impose a maximum number of updates in the subsolver, so that it does not spend too much time on the subproblem.

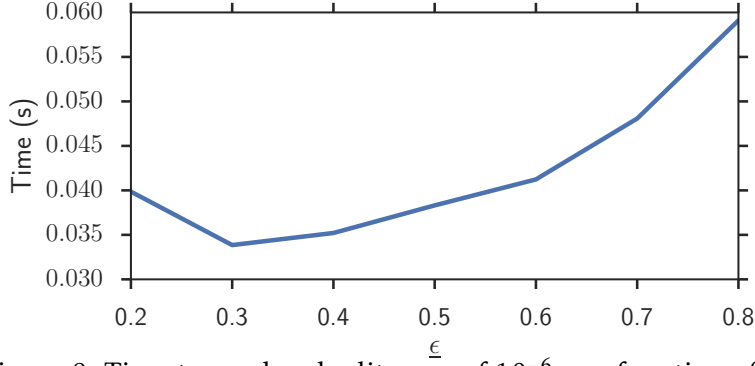

Figure 8: Time to reach a duality gap of $10^{-6}$ as a function of $\underline{\epsilon}$.

## D  Gap Safe screening in outer loop

Since the working sets in Algorithm 1 are constructed using the quantities $d_j(\Theta)$, it is a free byproduct of the WS definition to implement Gap Safe screening rules via Equation (2.2), since the only additional computation required is $\sqrt{\frac{2}{\lambda^2}\mathcal{G}^{(\lambda)}(\mathrm{B},\Theta)}$. Although Gap Safe screening is unlikely to exclude features which would be included in the WS (since Gap Safe rules discard features with $d_j(\Theta) > \sqrt{\frac{2}{\lambda^2}\mathcal{G}^{(\lambda)}(\mathrm{B},\Theta)}$ and the WS contains features with the lowest $d_j(\Theta)$) values, discarding features alleviates the computational cost of $\alpha_t$ and diminishes the number of $d_j(\Theta)$ to compute in the following outer loop iterations.

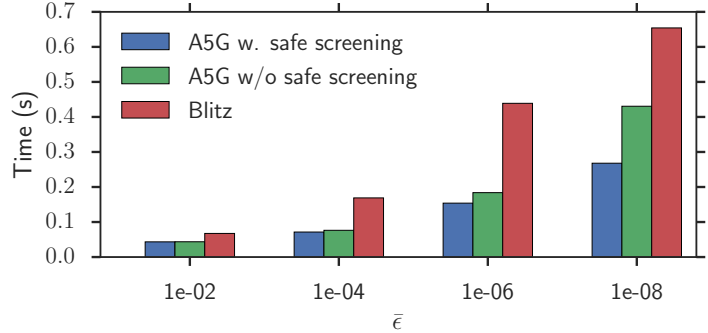

Figure 9: Computation time of a full Lasso path on the Leukemia dataset for various values of $\bar{\epsilon}$.

Figure 9 shows the benefits of combining safe screening with A5G when computing a Lasso path. The Lasso path is computed on 10 values of $\lambda$, chosen logarithmically between $\lambda_{max}/1000$ and $\lambda_{max}$. The solutions are computed up to a duality gap of $\bar{\epsilon}$, using the solution computed for the previous $\lambda$ as initialization. When safe screening is used, the fact that a solution for a close $\lambda$ is available restricts the number of features likely to integrate the WS, leading to a speed-up. When $\bar{\epsilon}$ is lower, the precision of the solution is better and safe screening gives larger speed-ups.